\documentclass{bmvc2k}

\title{Trajectory Space Factorization for Deep Video-Based 3D Human Pose Estimation}

\addauthor{Jiahao Lin}{jiahao@comp.nus.edu.sg}{1}
\addauthor{Gim Hee Lee}{gimhee.lee@comp.nus.edu.sg}{1}

\addinstitution{
 Department of Computer Science,\\
 School of Computing, \\
 National University of Singapore\\
}

\runninghead{Lin \etal{}}{Trajectory Space Factorization for 3D Human Pose Estimation}

\def\eg{\emph{e.g}\bmvaOneDot}

\def\etal{\emph{et al}\bmvaOneDot}

\usepackage{amsfonts}
\usepackage{commath}
\usepackage[caption=false]{subfig}
\usepackage{booktabs}
\begin{document}

\maketitle


\begin{abstract}

Existing deep learning approaches on 3d human pose estimation for videos are either based on Recurrent or Convolutional Neural Networks (RNNs or CNNs). However, RNN-based frameworks can only tackle sequences with limited frames because sequential models are sensitive to bad frames and tend to drift over long sequences. Although existing CNN-based temporal frameworks attempt to address the sensitivity and drift problems by concurrently processing all input frames in the sequence, the existing state-of-the-art CNN-based framework is limited to 3d pose estimation of a single frame from a sequential input.
In this paper, 
we propose a deep learning-based framework that utilizes matrix factorization for sequential 3d human poses estimation.  Our approach processes all input frames concurrently to avoid the sensitivity and drift problems, and yet outputs the 3d pose estimates for every frame in the input sequence.
More specifically, the 3d poses in all frames are represented as a motion matrix factorized into a trajectory bases matrix and a trajectory coefficient matrix.  
The trajectory bases matrix is precomputed from matrix factorization approaches such as 
Singular Value Decomposition (SVD) or Discrete Cosine Transform (DCT), and the problem of sequential 3d pose estimation is reduced to training a deep network to regress the trajectory coefficient matrix. 
We demonstrate the effectiveness of our framework on long sequences by achieving state-of-the-art performances on multiple benchmark datasets.
Our source code is available at: \url{https://github.com/jiahaoLjh/trajectory-pose-3d}.

\end{abstract}


\section{Introduction}
\label{sec:intro}

3d human pose estimation from a 2d image is an important task in computer vision that has a wide range of applications such as human-computer interaction, augmented/virtual reality, camera surveillance, \textit{etc}.
Nonetheless, 3d pose estimation remains a challenging problem due to the inherent ambiguity in the ill-posed problem of lifting 3d pose from a 2d image.

Over the years, numerous research has been done on 3d pose estimation from a single image and a sequence of input frames, respectively. 
Despite the intuition that 3d pose estimation should be improved by having more information from the sequence of input frames, the performance of temporal-based frameworks are unexpectedly limited compared to its single frame counterpart. One of the reasons is sequential frameworks are mostly based on RNNs \cite{hossain2018exploiting, lin2017recurrent, lee2018propagating} that are very sensitive to the quality of the estimate at each time step. Just a few frames of bad estimates would lead to dire results in the estimation for all subsequent frames. 
Recently, CNN-based temporal frameworks \cite{collobert2016wav2letter, dauphin2017language, gehring2017convolutional, van2016wavenet} have shown advantages over RNNs in modeling temporal information. However, concurrently estimating all frames in a long sequence is an arduous task for data-driven approaches due to the increasing dimensionality of the output space with longer sequences, and the network for ``many-to-many mapping" requires significantly more data to train. As a result, the state-of-the-art temporal framework for 3d pose estimation \cite{pavllo20183d} only outputs the estimate of a single frame centered on an input sequence of a few hundred frames.

A key observation is that consecutive frames of human motions are highly correlated and abrupt changes in the motions are unlikely. This suggests that we do not need to estimate the 3d pose in each frame of a sequence independently. Instead, we can effectively reduce the output dimension by decomposing the uncorrelated components. To this end, we propose a deep learning-based framework that utilizes matrix factorization \cite{tomasi1992shape, bregler2000recovering, akhter2009nonrigid} for sequential 3d human poses estimation.
More specifically, the joints of the 3d pose estimates in all frames are represented as a motion matrix factorized into a trajectory bases matrix and a trajectory coefficient matrix. The trajectory bases matrix is precomputed with 
SVD or cosine bases from DCT, and the problem of sequential 3d pose estimation is reduced to training a deep network to regress the trajectory coefficient matrix. 
We show that a relatively small number of bases (compared to the large number of input frames) can effectively approximate the human motions.
We transform the extracted features from Cartesian space to trajectory space and regress the trajectory coefficients with a densely connected multilayer perceptron (MLP). The 3d poses can then be reconstructed from the linear combination of the trajectory bases with the trajectory coefficients regressed from the deep network.

Our main contributions are: (1) we present an effective framework to estimate 3d human poses for a long sequence in the trajectory space; (2) we show experimentally that a small number of trajectory bases are sufficient to model human motions which greatly reduces the complexity of the network training; and (3) our method gives state-of-the-art performance on multiple benchmark datasets.


\section{Related Work}
\label{sec:literature}
Early methods for
3d human pose estimation from 2d images
use a variety of hand-crafted features such as silhouette \cite{agarwal20043d}, shape \cite{mori2006recovering}, SIFT \cite{bo2008fast}, HOG \cite{rogez2008randomized} to estimate 3d human poses. More recently, a large number of works are based on the highly popular and effective deep neural networks. 
Li \etal \cite{li20143d} apply a deep convolutional neural network to jointly regress pose and detect body parts.
Tekin \etal \cite{tekin2016structured} propose to learn a high-dimensional latent pose representation accouting for joint dependencies using auto-encoder.
Pavlakos \etal \cite{pavlakos2017coarse} extend the idea of regressing heat-map from 2d to 3d and use a coarse-to-fine approach to estimate the 3d heat-map for joint estimation.
Sun \etal \cite{sun2017compositional} exploit the skeleton structure and define a compositional loss function to better learn joint interactions.
Lee \etal \cite{lee2018propagating} propose a structure-aware multi-stage architecture to regress body parts incrementally.
Training deep learning models directly from images requires huge amounts of data to be labeled with 3d ground truth. To alleviate this problem, a two-stage framework is proposed in several works. In particular, 2d poses are first estimated and subsequently used for 3d pose estimation.
Yasin \etal \cite{yasin2016dual} and Chen \etal \cite{chen20173d} exploit the idea of 3d pose retrieval by retrieving the nearest 3d poses in a 3d library based on the 2d poses.
Moreno-Noguer \cite{moreno20173d} estimates the 3d pose by regressing a distance matrix from 2d space to 3d space.
Martinez \etal \cite{martinez2017simple} propose a simple yet effective regression network based on fully-connected residual blocks.
Fang \etal \cite{fang2018learning} integrate kinematic knowledge into their framework and design a hierarchical grammar model.
Zhao \etal \cite{zhao2019semantic} propose to use graph neural networks to model structural information in human poses.
Our work also follows the two-stage approach.

The inherent depth ambiguity in 3d pose estimation from monocular images limits the estimation accuracy. Extensive research has been done to exploit extra information contained in temporal sequences.
Zhou \etal \cite{zhou2016sparseness} formulate an optimization problem to search for the 3d configuration with the highest probability given 2d confidence maps and solve the problem using Expectation-Maximization.
Tekin \etal \cite{tekin2016direct} use a CNN to align bounding box of consecutive frames and then generate a spatial-temporal volume based on which they extract 3d HOG features and regress the 3d pose for the central frame.
Mehta \etal \cite{mehta2017vnect} propose a real-time system for 3d pose estimation and apply temporal filtering to yield temporally consistent 3d poses.

Recently, RNN-based frameworks are used to deal with sequential input data.
Lin \etal \cite{lin2017recurrent} use a multi-stage framework based on Long Short-term Memory (LSTM) units to estimate the 3d pose from the extracted 2d features and estimated 3d pose in the previous stage.
Coskun \etal \cite{coskun2017long} propose to learn a human motion model using Kalman Filter and implement it with LSTMs.
Hossain \etal \cite{hossain2018exploiting} design a sequence-to-sequence network with LSTM units to first encode a sequence of motions in the form of 2d joint locations and then decode the 3d poses of the sequence.
However, RNNs are sensitive to erroneous inputs and tend to drift over long sequences.
To overcome the shortcomings of RNNs, a CNN-based framework is proposed by Pavllo \etal \cite{pavllo20183d} to aggregate temporal information using dilated convolutions. Despite being successful at regressing a single frame from a sequence of input, it cannot concurrently output the 3d pose estimations for all frames in the sequence. 

Inspired by matrix factorization methods commonly used in Structure-from-Motion (SfM) \cite{tomasi1992shape} and non-rigid SfM \cite{bregler2000recovering}, several works \cite{ramakrishna2012reconstructing, zhou20153d, zhou2016sparseness} on 3d human pose estimation factorize the sequence of 3d human poses into a linear combination of shape bases. Akhter \etal \cite{akhter2009nonrigid} suggest a duality of the factorization in the trajectory space. We extend the idea of 
matrix factorization to learning a deep network that estimates the coefficients of the trajectory bases from a sequence of 2d poses as inputs. The 3d poses of all frames are recovered concurrently as the linear combinations of the trajectory bases with the estimated coefficients.  


\section{Our approach}
\label{sec:approach}

In this section, we introduce our framework for sequential 3d pose estimation in detail. We formulate the estimation problem as a trajectory space coefficient regression task (Sec \ref{sec:factorization}). In particular, we use fixed trajectory bases, e.g., singular vectors on motion data or DCT bases, as the trajectory bases (Sec \ref{sec:trajectory-bases}), and propose a novel network architecture to effectively regress the trajectory coefficients from a sequence of 2d input (Sec \ref{sec:network}). An overview of our framework is shown in Figure \ref{fig:network}.

\subsection{Non-Rigid Structure Factorization}
\label{sec:factorization}

Given the 2d image coordinates $(u,v)$ of $J$ joints in $F$ consecutive frames, the goal is to estimate the 3d Cartesian coordinates $(X,Y,Z)$ of the $J$ joints in all $F$ frames. Let us denote the 3d coordinates of the $J$ joints in the $f$-th frame as
\begin{equation}
    S_f = \begin{bmatrix}
                X_{f1} & Y_{f1} & Z_{f1} & \dots & X_{fJ} & Y_{fJ} & Z_{fJ}
            \end{bmatrix}. 
\end{equation}
The entire sequence of $F$ frames are concatenated to form the motion matrix,
\begin{equation}
    \mathbf{S}_{F \times 3J} = \begin{bmatrix}
                X_{11} & Y_{11} & Z_{11} & \dots & X_{1J} & Y_{1J} & Z_{1J} \\
                \vdots & \vdots & \vdots & & \vdots & \vdots & \vdots \\
                X_{F1} & Y_{F1} & Z_{F1} & \dots & X_{FJ} & Y_{FJ} & Z_{FJ}
        \end{bmatrix},
\end{equation}
where the column space is known as the \textit{trajectory space} \cite{akhter2009nonrigid}. Furthermore, $\mathbf{S}$ can be factorized into a linear combination of $K$ trajectory bases, i.e., 
\begin{equation}
\label{equ: factorization}
    \mathbf{S}_{F \times 3J} = \underbrace{\begin{bmatrix}
        \theta_{1}^{(1)} & \dots & \theta_{1}^{(K)} \\
        \vdots & & \vdots \\
        \theta_{F}^{(1)} & \dots & \theta_{F}^{(K)}
    \end{bmatrix}_{F \times K}}_{\Theta_{F \times K}} \times \underbrace{\begin{bmatrix}
        a_{x1}^{(1)} & a_{y1}^{(1)} & a_{z1}^{(1)} & \dots & a_{xJ}^{(1)} & a_{yJ}^{(1)} & a_{zJ}^{(1)} \\
        \vdots & \vdots & \vdots & & \vdots & \vdots & \vdots \\
        a_{x1}^{(K)} & a_{y1}^{(K)} & a_{z1}^{(K)} & \dots & a_{xJ}^{(K)} & a_{yJ}^{(K)} & a_{zJ}^{(K)}
    \end{bmatrix}_{K \times 3J}}_{\mathbf{A}_{K \times 3J}}.
\end{equation}
$\Theta_{F \times K}$ is the trajectory bases matrix, where each column $[\begin{smallmatrix} \theta_{1}^{(k)} & \dots & \theta_{F}^{(k)} \end{smallmatrix}]^{T} \in \mathbb{R}^{F}$ is a trajectory basis vector.
Here, the number of trajectory bases $K$ is no greater than the number of frames $F$ since $\text{rank}(\mathbf{S}_{F \times 3J}) \le \min(F, 3J)$. Given a set of predefined trajectory bases $\Theta$, the estimation of a sequence of 3d poses $\mathbf{S}$ turns into the problem of finding the coefficient matrix $\mathbf{A}$.
On the other hand, the row space of $\mathbf{S}$ is known as the \textit{shape space}, where
the factorization in Equation \ref{equ: factorization} can also be interpreted as the product of a coefficient matrix $\Theta$ in the shape space and a shape bases matrix $\mathbf{A}$. The duality of the factorization in the row and column spaces is called \textit{shape-trajectory duality}. 
Unlike approaches \cite{ramakrishna2012reconstructing, zhou20153d, zhou2016sparseness} that estimate the 3d poses in shape space, we follow \cite{akhter2009nonrigid} to formulate the task in the trajectory space. This is because shape bases are highly motion specific, hence, large number of bases are required to achieve low reconstruction error for any arbitrary motion type \cite{ramakrishna2012reconstructing}. In contrast, trajectory space requires significantly fewer dominant bases to model human motions due to the physics constraining the motion acceleration. Additionally, reconstructed motions from linear combination of smooth trajectory bases are temporally consistent.

\subsection{Trajectory Bases}
\label{sec:trajectory-bases}

\begin{figure*}[t]
\centering
\subfloat[]{\includegraphics[width=0.48\linewidth]{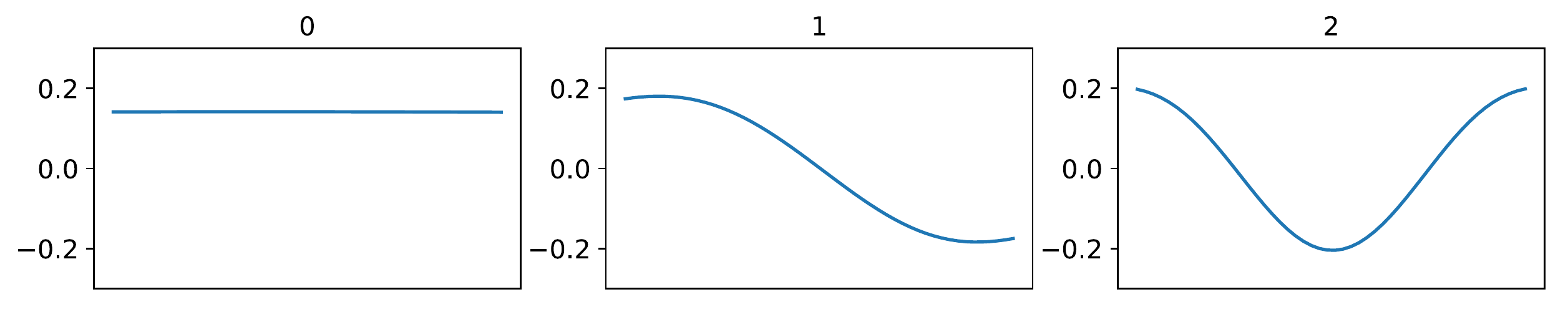}\label{fig:svd}}
\quad
\subfloat[]{\includegraphics[width=0.48\linewidth]{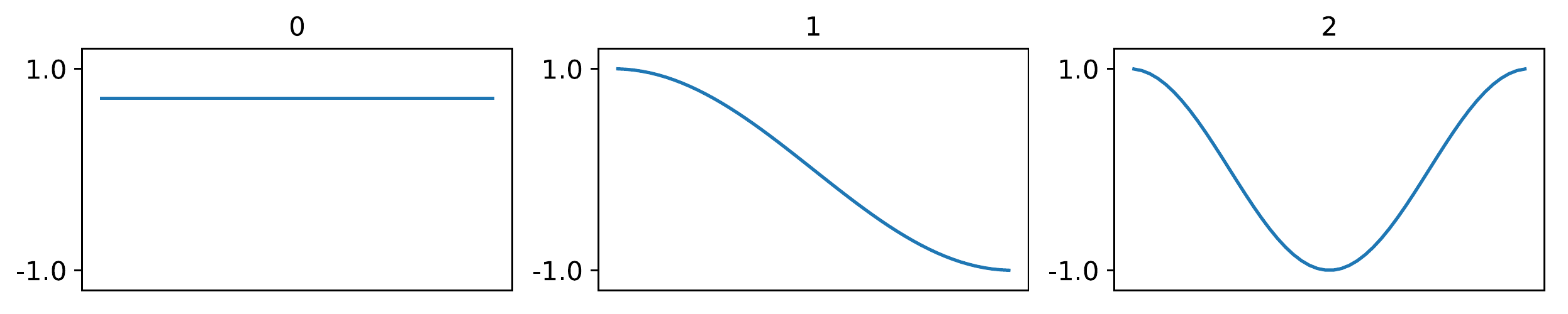}\label{fig:dct}}
\caption{Trajectory bases for $F=50$ frames. \textbf{(a)}: Singular vectors corresponding to the 3 largest singular values as trajectory bases. \textbf{(b)}: First 3 bases from DCT.}
\label{fig:bases}
\end{figure*}

\paragraph{Singular Value Decomposition (SVD).} The trajectory bases matrix $\Theta_{F \times K}$ can be computed from the SVD of the motion matrix $\mathbf{S}$, i.e., $\mathbf{S}_{F \times 3J} = U_{F \times F} \Sigma_{F \times 3J} V_{3J \times 3J}$,
where $U_{F \times F}$ and $V_{3J \times 3J}$ are the left and right singular vectors, and $\Sigma_{F \times 3J}$ is the diagonal matrix of singular values. 
More specifically, $\Theta_{F \times K}$ is the $K$ columns of the left singular vectors $U_{F \times F}$ that correspond to the $K$ largest singular values.   
Multiple motion matrices are stacked into a larger matrix 
$\Tilde{\mathbf{S}}_{F \times (N \times 3J)}$, where $N$ denotes the total number of motion matrices. The same SVD operation is applied on $\Tilde{\mathbf{S}}_{F \times (N \times 3J)}$ to get $\Theta_{F \times K}$. A higher number of motion matrices, i.e., large $N$, results in a more accurate trajectory bases matrix $\Theta_{F \times K}$. However, in practice
we are limited by the memory needed to compute SVD. Hence, we
compute $\Theta_{F \times K}$ from a small portion of the available large scale datasets \eg Human3.6M \cite{ionescu2014human3}. Nonetheless, the result is stable enough to show the trajectory pattern within human motion. 
Figure \ref{fig:svd} shows the plots of the bases computed by SVD on 10k trajectories from Human3.6M. 
Each plot is the $F$ entries of a trajectory basis vector $\theta^{(k)}$ (y-axis) against its index (x-axis) in the vector.  

\begin{figure*}[t]
\includegraphics[width=\linewidth]{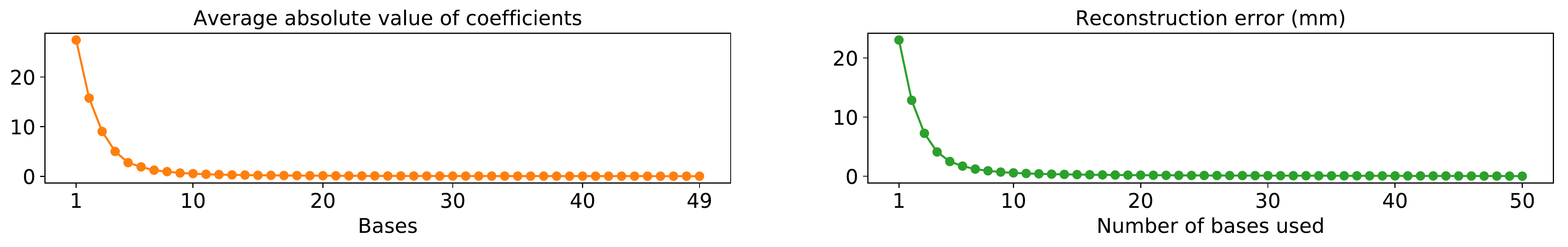}
\caption{\textbf{Left:} Mean of absolute coefficient values corresponding to different DCT bases. The first coefficient corresponding to the DC component of a signal is discarded in the figure. \textbf{Right:} Reconstruction error when truncated to different number of DCT bases.}
\label{fig:truncate}
\end{figure*}

\paragraph{Discrete Cosine Transform (DCT).} The plots of the trajectory bases vectors from Figure \ref{fig:svd} show strong indication of sinusoidal form. Based on this observation, we also explore DCT that compactly model motion trajectories as pointed out in \cite{akhter2009nonrigid}.
For a discrete signal input $\{d_0, d_1, \dots, d_{F-1}\}$ of finite length $F$, the DCT is defined as
\begin{equation}
    D_k = \sum_{f=0}^{F-1}d_f\cos{\Big[\frac{\pi}{F}\big(f+\frac{1}{2}\big)k\Big]}, \hspace{5mm} k=0,\dots,F-1.
\label{equ:dct1}
\end{equation}
$D_k$ stands for the coefficient corresponding to the $k$-th cosine wave $\cos{[\frac{\pi}{F}(f+\frac{1}{2})k]}$. The inverse DCT is used to transform the coefficients back to the original signal, i.e., 
\begin{equation}
    d_f = \frac{1}{2}D_0 + \sum_{k=1}^{F-1}D_k\cos{\Big[\frac{\pi}{F}\big(f+\frac{1}{2}\big)k\Big]}, \hspace{5mm} f=0,\dots,F-1.
\label{equ:dct2}
\end{equation}
Furthermore, the inverse DCT can be seen as a linear combination of the orthogonal bases $0.5,~k=0$ and $\cos{[\frac{\pi}{F}(f+\frac{1}{2})k]},~k=1,\dots,F-1$ with the coefficients $\{D_0, \dots, D_{F-1}\}$, and $[d_0, \dots, d_{F-1}]^T$ is a column in the motion matrix $\mathbf{S}$. Limiting $k=0, \dots, K-1$ gives us a set of $F \times K$ orthogonal bases that is similar to $\Theta_{F \times K}$ mentioned earlier. The advantage of DCT is that the orthogonal bases are predefined without the need to learn from data. Figure \ref{fig:dct} shows the first 3 bases of DCT, which resemble the singular vectors in Figure \ref{fig:svd}.

Figure \ref{fig:truncate} (left) shows an example analysis on a subset of Human3.6M dataset (randomly sampling 100k trajectories with $F=50$). We see that bases with $k>10$ have negligible corresponding coefficients, which is reasonable because human motions usually contain very sparse high-frequency movement. Figure \ref{fig:truncate} (right) further shows that it is sufficient to reconstruct the trajectory with relatively low error even if a very small proportion of bases are utilized. Hence, we only need to regress a low number of coefficients despite large number of frames because the trajectory bases already encodes the temporal correlation within the motion. This greatly reduces the complexity of training deep networks.

\subsection{Network Design and Implementation Details}
\label{sec:network}

\begin{figure*}
\includegraphics[width=\linewidth]{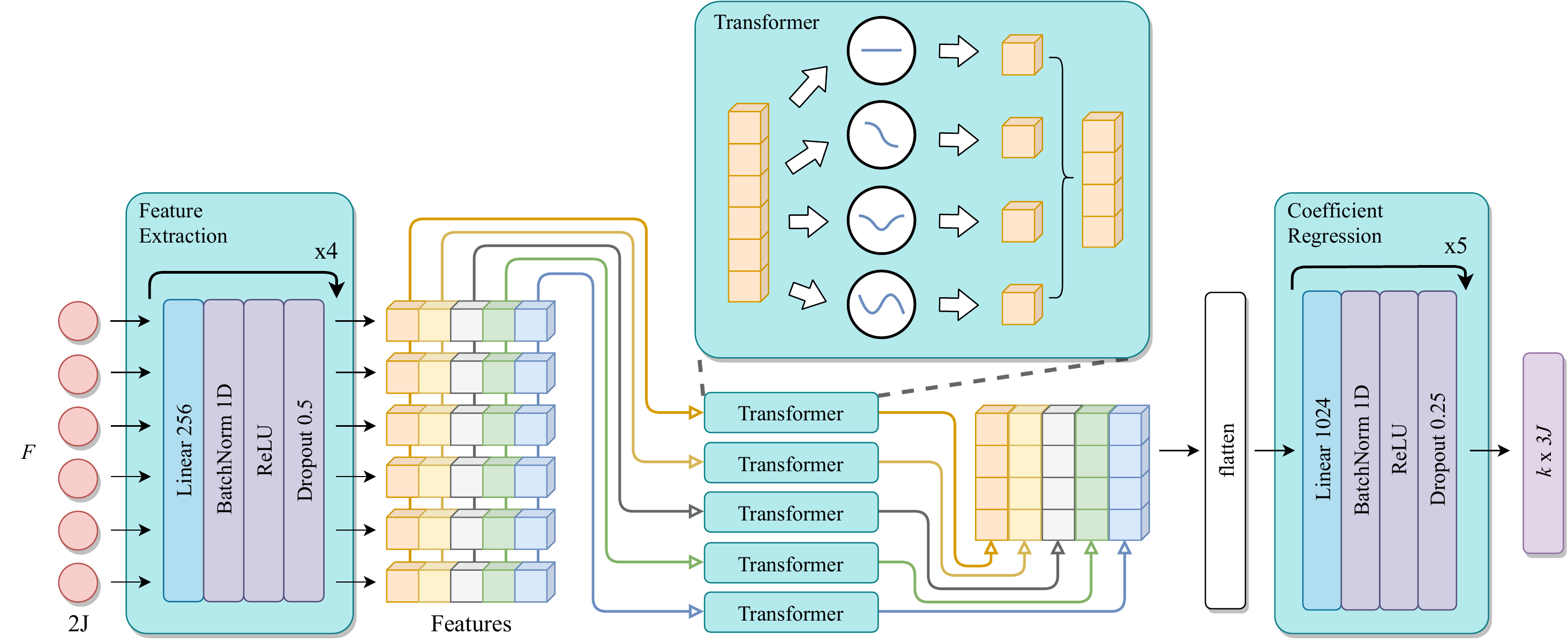}
\caption{Our Network Architecture. $F$ frames of $J$ 2d joints are fed into a MLP for per frame feature extraction. Each feature channel along the temporal axis is transformed into trajectory space via a \textit{Transformer}. Coefficients from all feature channels are then concatenated and another MLP is applied to regress the $K$ coefficients for all $3J$ trajectories.}
\label{fig:network}
\end{figure*}

Figure \ref{fig:network} shows an overview of our network. The input of the network is the 2d image coordinates of $J$ joints from a sequence of $F$ consecutive frames obtained from any off-the-shelf 2d pose estimators, \eg \cite{newell2016stacked, chen2018cascaded}. A MLP block with 4 repeated \textit{Linear-BatchNorm-ReLU-Dropout} structures is used to extract features for each frame. The weights within the block are shared across all frames. Each feature channel across the temporal axis is stacked into a "\textit{virtual feature trajectory}" and is transformed into the trajectory space via a "\textit{Transformer}". The "\textit{Transformer}" is simply an implementation of the DCT. An inner product is computed for the input trajectory and each of the fixed trajectory bases, and is further normalized by multiplying a scale factor of $\frac{2}{F}$. This results in $K$ coefficients for each feature channel, and all coefficients from all channels are concatenated and passed into another MLP block to regress the final $K$ coefficients for all the $J \times 3$ trajectories ($X,Y,Z$ coordinates of $J$ joints). The "\textit{Coefficient Regression}" MLP block consists of 5 repeated \textit{Linear-BatchNorm-ReLU-Dropout} structures, and differs from the "\textit{Feature Extraction}" block for both the linear layer size and dropout rate. Finally, we linearly combine the $K$ trajectory bases using the regressed coefficients to reconstruct the $X,Y,Z$ values for all $J$ joints across $F$ frames.

We empirically found that by densely connecting any two linear layers in the MLP block (originally proposed for convolutional layers by \cite{huang2017densely}), the network is capable of learning the length of the critical path automatically. This helps to boost the regression performance. The skip connection arrow in both "\textit{Feature Extraction}" block and "\textit{Coefficient Regression}" block indicates the usage of dense shortcut links.
Temporal-based frameworks usually suffer from noisy 2d input. To alleviate the detrimental impact from individual noisy input frames, an extra average pooling layer is added after feature extraction block to smoothen the "feature trajectories". We found that a pooling window of length 5 is suitable to reduce noise while preserving much of the underlying signal, and we report all the experiments results under this setting.
Our network is not restricted to any particular input sequence length. It can be adapted to any sequence length by simply substituting the trajectory bases in the "\textit{Transformer}". The number of trajectory bases can also be adjusted by adding (or removing) branches in the "\textit{Transformer}". Experiments on how the number of input frames and the number of trajectory bases affect the performance are shown in Section \ref{sec:result}.

We supervise the training of our network by minimizing \textit{L}1-loss over all $N$ sequences
\begin{equation}
    \mathcal{L}_{3D}(\hat{\mathbf{S}})=\frac{1}{N}\frac{1}{F}\sum_{i=1}^{N}\lVert \hat{\mathbf{S}}_i - \mathbf{S}_i \rVert_1
\end{equation}
because \textit{L}1-loss is more robust to outliers which are common due to occlusions. Here $\hat{\mathbf{S}}_i$ denotes the estimated 3d poses while $\mathbf{S}_i$ denotes the corresponding ground truth poses. We train our model using Adam \cite{kingma2014adam} optimizer for 100 epochs with initial learning rate of $1e-4$. We schedule the learning rate decay at epoch 60 and 85 with a shrink factor $\alpha=0.1$. We also perform data augmentation by horizontally flipping the human pose. During inference, estimated 3d poses for both original and flipped data are averaged as the final estimation.


\section{Experiments}
\label{sec:result}

\paragraph{Experimental setup.} We focus our evaluation on two widely used datasets: Human3.6M \cite{ionescu2014human3} and MPI-INF-3DHP \cite{mehta2016monocular}. We show both quantitative and qualitative results that demonstrate the effectiveness of our temporal estimation framework. \textbf{Human3.6M} is currently the largest publicly available dataset for human 3d pose estimation. The dataset consists of 3.6 million image frames captured by MoCap system in a constrained indoor studio environment. 11 actors perform 15 everyday activities such as walking, discussing, \textit{etc}. We follow two standard protocols for comparison with previous works. We use 5 subjects (S1, S5, S6, S7, S8) for training and 2 subjects (S9, S11) for testing. The evaluation metric is mean per joint position error (MPJPE) in mm after aligning the hip joint to the origin. We refer to this as {\bf protocol 1}. Several works \cite{bogo2016keep,pavlakos2017coarse,sun2017compositional,lee2018propagating,dabral2018learning,zhou2016sparseness,moreno20173d,nie2017monocular,martinez2017simple,fang2018learning,hossain2018exploiting, pavllo20183d} also report the error after aligning further with respect to the ground truth pose via Procrustes Analysis. We refer to this as {\bf protocol 2}. \textbf{MPI-INF-3DHP} is a recently released 3d dataset with both indoor environment (green background and studio background) and in-the-wild outdoor environment. Similar to \cite{mehta2016monocular}, we report the 3D Percentage of Correct Keypoints (PCK) with threshold of 150mm and Area Under Curve (AUC) for a range of PCK thresholds.

We follow \cite{pavllo20183d} in using 2d detections from the \textit{Cascaded Pyramid Network (CPN)} \cite{chen2018cascaded} as our network input. Instead of down-sampling, which is commonly done in many works that use a single-frame input, we keep the original frame rate (50fps for Human3.6M and 25fps for MPI-INF-3DHP) because having access to the complete sequence provides more detailed information. 
The trajectory bases are the SVD or DCT bases described in Section \ref{sec:trajectory-bases}
for any predefined number of input frames $F$. More specifically, we perform SVD on the motion matrix formed with 10k randomly sampled trajectories from the training set to get the SVD bases. The DCT bases are the cosine bases defined in Equation \ref{equ:dct1} and \ref{equ:dct2}. 
We train the network with a fixed number of $K$ trajectory bases and $F$ input frames. Nonetheless, we use the ``sliding window approach" on input videos with $L > F$ frames. In particular, we infer the 3d poses on a sliding window of $F$ frames 
for an input video with an arbitrary $L > F$ frames. We move the sliding window at a step of $q < F$ frames, where $q$ is set to 5 for all experiments shown in this paper.
Since the sliding window runs through most of the frames multiple times (except for the first and last frames), 
we compute the final pose for each frame as the average of all the 3d poses estimated for that frame. 
We note that $L$ can range from 300 to 6000 frames for all the videos of the two datasets used in our experiments.

\renewcommand{\arraystretch}{1.1}

\begin{table*}[t]
    \centering
    \resizebox{\linewidth}{!}{
        \begin{tabular}{l|rrrrrrrrrrrrrrr|r}
            \hline
            Protocol 1 & Direct. & Discuss & Eating & Greet & Phone & Photo & Pose & Purch. & Sitting & SitingD & Smoke & Wait & WalkD & Walk & WalkT & Avg \\
            
            \hline \hline
            
            Pavlakos \etal \cite{pavlakos2017coarse} & 67.4 & 71.9 & 66.7 & 69.1 & 72.0 & 77.0 & 65.0 & 68.3 & 83.7 & 96.5 & 71.7 & 65.8 & 74.9 & 59.1 & 63.2 & 71.9 \\
            Tekin \etal \cite{tekin2017learning} & 54.2 & 61.4 & 60.2 & 61.2 & 79.4 & 78.3 & 63.1 & 81.6 & 70.1 & 107.3 & 69.3 & 70.3 & 74.3 & 51.8 & 63.2 & 69.7\\
            Martinez \etal \cite{martinez2017simple} & 51.8 & 56.2 & 58.1 & 59.0 & 69.5 & 78.4 & 55.2 & 58.1 & 74.0 & 94.6 & 62.3 & 59.1 & 65.1 & 49.5 & 52.4 & 62.9\\
            Fang \etal \cite{fang2018learning} & 50.1 & 54.3 & 57.0 & 57.1 & 66.6 & 73.3 & 53.4 & 55.7 & 72.8 & 88.6 & 60.3 & 57.7 & 62.7 & 47.5 & 50.6 & 60.4\\
            Sun \etal \cite{sun2017compositional} & 52.8 & 54.8 & 54.2 & 54.3 & 61.8 & 67.2 & 53.1 & 53.6 & 71.7 & 86.7 & 61.5 & 53.4 & 61.6 & 47.1 & 53.4 & 59.1\\
            Yang \etal \cite{yang20183d} & 51.5 & 58.9 & 50.4 & 57.0 & 62.1 & 65.4 & 49.8 & 52.7 & 69.2 & 85.2 & 57.4 & 58.4 & 60.1 & 43.6 & 47.7 & 58.6\\
            Zhao \etal \cite{zhao2019semantic} & 47.3 & 60.7 & 51.4 & 60.5 & 61.1 & 67.8 & 49.9 & 47.3 & 68.1 & 86.2 & 55.0 & 61.0 & 60.6 & 42.1 & 45.3 & 57.6\\
            Lee \etal \cite{lee2018propagating} ($F=1$) & 43.8 & 51.7 & 48.8 & 53.1 & 52.2 & 74.9 & 52.7 & 44.6 & 56.9 & 74.3 & 56.7 & 66.4 & 68.4 & 47.5 & 45.6 & 55.8\\
            Dabral \etal \cite{dabral2018learning} (SAP-Net) & 46.9 & 53.8 & 47.0 & 52.8 & 56.9 & 63.6 & 45.2 & 48.2 & 68.0 & 94.0 & 55.7 & 51.6 & 55.4 & 40.3 & 44.3 & 55.5\\
            
            \hline \hline
            
            Zhou \etal \cite{zhou2016sparseness} & 87.4 & 109.3 & 87.1 & 103.2 & 116.2 & 143.3 & 106.9 & 99.8 & 124.5 & 199.2 & 107.4 & 118.1 & 114.2 & 79.4 & 97.7 & 113.0\\
            Lin \etal \cite{lin2017recurrent} & 58.0 & 68.2 & 63.3 & 65.8 & 75.3 & 93.1 & 61.2 & 65.7 & 98.7 & 127.7 & 70.4 & 68.2 & 72.9 & 50.6 & 57.7 & 73.1\\
            Hossain \& Little \cite{hossain2018exploiting} & 48.4 & 50.7 & 57.2 & 55.2 & 63.1 & 72.6 & 53.0 & 51.7 & 66.1 & 80.9 & 59.0 & 57.3 & 62.4 & 46.6 & 49.6 & 58.3\\
            Lee \etal \cite{lee2018propagating} ($F=3$) & \textbf{40.2} & 49.2 & 47.8 & 52.6 & 50.1 & 75.0 & 50.2 & 43.0 & \textbf{55.8} & 73.9 & 54.1 & 55.6 & 58.2 & 43.3 & 43.3 & 52.8\\
            Dabral \etal \cite{dabral2018learning} (TP-Net) & 44.8 & 50.4 & 44.7 & 49.0 & 52.9 & 61.4 & 43.5 & 45.5 & 63.1 & 87.3 & 51.7 & 48.5 & 52.2 & 37.6 & 41.9 & 52.1\\
            Pavllo \etal \cite{pavllo20183d} & 45.2 & 46.7 & 43.3 & 45.6 & \textbf{48.1} & \textbf{55.1} & 44.6 & 44.3 & 57.3 & 65.8 & \textbf{47.1} & 44.0 & 49.0 & \textbf{32.8} & \textbf{33.9} & 46.8\\
            
            Ours ($F=10$) & 43.2 & 46.7 & 45.5 & 46.1 & 51.5 & 59.2 & 44.5 & 42.9 & 58.0 & 66.2 & 49.2 & 45.5 & 50.3 & 36.5 & 39.6 & 48.8\\
            Ours ($F=25$) & 42.7 & 45.5 & 43.1 & 44.6 & 49.5 & 57.3 & 43.1 & 41.7 & 57.5 & 65.0 & 48.0 & 43.6 & 48.5 & 34.2 & 36.9 & 47.3\\
            Ours ($F=50$) & 42.5 & \textbf{44.8} & \textbf{42.6} & \textbf{44.2} & 48.5 & 57.1 & \textbf{42.6} & \textbf{41.4} & 56.5 & \textbf{64.5} & 47.4 & \textbf{43.0} & \textbf{48.1} & 33.0 & 35.1 & \textbf{46.6}\\
            \hline
        \end{tabular}
    }
    \caption{Results on Human3.6M under Protocol 1 (no rigid alignment for post-processing). Top half of the table are single-frame works. Bottom half of the table are multi-frame works. Bold-faced numbers indicate best results.}
    \label{tab:protocol1}
\end{table*}

\begin{table*}[t]
    \centering
    \resizebox{\linewidth}{!}{
        \begin{tabular}{l|rrrrrrrrrrrrrrr|r}
            \hline
            Protocol 2 & Direct. & Discuss & Eating & Greet & Phone & Photo & Pose & Purch. & Sitting & SitingD & Smoke & Wait & WalkD & Walk & WalkT & Avg \\
            
            \hline \hline
            
            Pavlakos \etal \cite{pavlakos2017coarse} & - & - & - & - & - & - & - & - & - & - & - & - & - & - & - & 51.9 \\
            Sun \etal \cite{sun2017compositional} & 42.1 & 44.3 & 45.0 & 45.4 & 51.5 & 53.0 & 43.2 & 41.3 & 59.3 & 73.3 & 51.0 & 44.0 & 48.0 & 38.3 & 44.8 & 48.3\\
            Martinez \etal \cite{martinez2017simple} & 39.5 & 43.2 & 46.4 & 47.0 & 51.0 & 56.0 & 41.4 & 40.6 & 56.5 & 69.4 & 49.2 & 45.0 & 49.5 & 38.0 & 43.1 & 47.7\\
            Lee \etal \cite{lee2018propagating} ($F=1$) & 38.0 & 39.3 & 46.3 & 44.4 & 49.0 & 55.1 & 40.2 & 41.1 & 53.2 & 68.9 & 51.0 & 39.1 & 56.4 & 33.9 & 38.5 & 46.2\\
            Fang \etal \cite{fang2018learning} & 38.2 & 41.7 & 43.7 & 44.9 & 48.5 & 55.3 & 40.2 & 38.2 & 54.5 & 64.4 & 47.2 & 44.3 & 47.3 & 36.7 & 41.7 & 45.7\\
            Dabral \etal \cite{dabral2018learning} (SAP-Net) & 32.8 & 36.8 & 42.5 & 38.5 & 42.4 & 49.0 & 35.4 & 34.3 & 53.6 & 66.2 & 46.5 & 34.1 & 42.3 & 30.0 & 39.7 & 42.2\\
            
            \hline \hline
            
            Hossain \& Little \cite{hossain2018exploiting} & 35.7 & 39.3 & 44.6 & 43.0 & 47.2 & 54.0 & 38.3 & 37.5 & 51.6 & 61.3 & 46.5 & 41.4 & 47.3 & 34.2 & 39.4 & 44.1\\
            Pavllo \etal \cite{pavllo20183d} & 34.1 & 36.1 & 34.4 & 37.2 & \textbf{36.4} & \textbf{42.2} & 34.4 & 33.6 & 45.0 & 52.5 & \textbf{37.4} & 33.8 & 37.8 & 25.6 & \textbf{27.3} & 36.5\\
            Dabral \etal \cite{dabral2018learning} (TP-Net) & \textbf{28.0} & \textbf{30.7} & 39.1 & \textbf{34.4} & 37.1 & 44.8 & \textbf{28.9} & \textbf{31.2} & \textbf{39.3} & 60.6 & 39.3 & \textbf{31.1} & 37.8 & \textbf{25.3} & 28.4 & \textbf{36.3}\\
            
            Ours ($F=10$) & 32.9 & 36.4 & 35.8 & 37.3 & 39.2 & 44.2 & 34.0 & 33.0 & 46.5 & 53.8 & 39.7 & 34.3 & 40.0 & 27.8 & 32.8 & 38.3\\
            Ours ($F=25$) & 32.8 & 35.7 & 34.4 & 36.1 & 38.1 & 43.3 & 33.0 & 32.5 & 46.4 & 52.7 & 38.7 & 33.1 & 38.5 & 26.3 & 30.7 & 37.3\\
            Ours ($F=50$) & 32.5 & 35.3 & \textbf{34.3} & 36.2 & 37.8 & 43.0 & 33.0 & 32.2 & 45.7 & \textbf{51.8} & 38.4 & 32.8 & \textbf{37.5} & 25.8 & 28.9 & 36.8\\
            \hline
        \end{tabular}
    }
    \caption{Results on Human3.6M under Protocol 2 (after rigid alignment for post-processing). Top half of the table are single-frame works. Bottom half of the table are multi-frame works. Bold-faced numbers indicate best results.}
    \label{tab:protocol2}
\end{table*}

\begin{figure*}[t]
\subfloat[]{\includegraphics[width=0.5\linewidth]{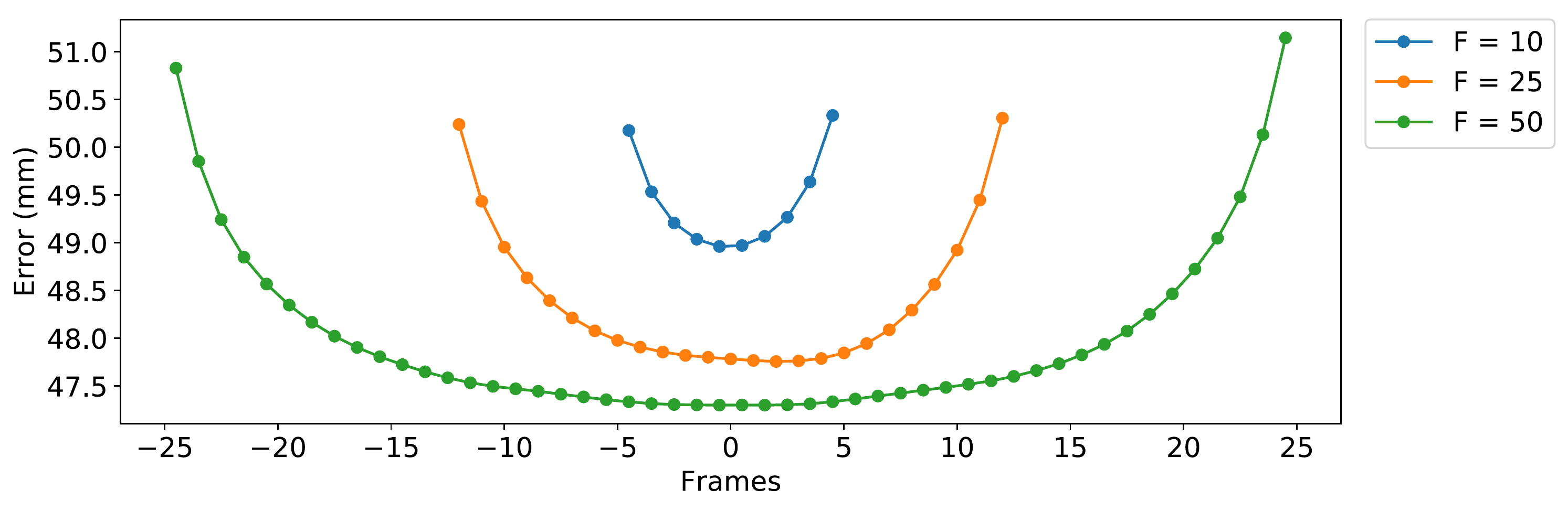}\label{fig:sequence}}
\subfloat[]{\includegraphics[width=0.5\linewidth]{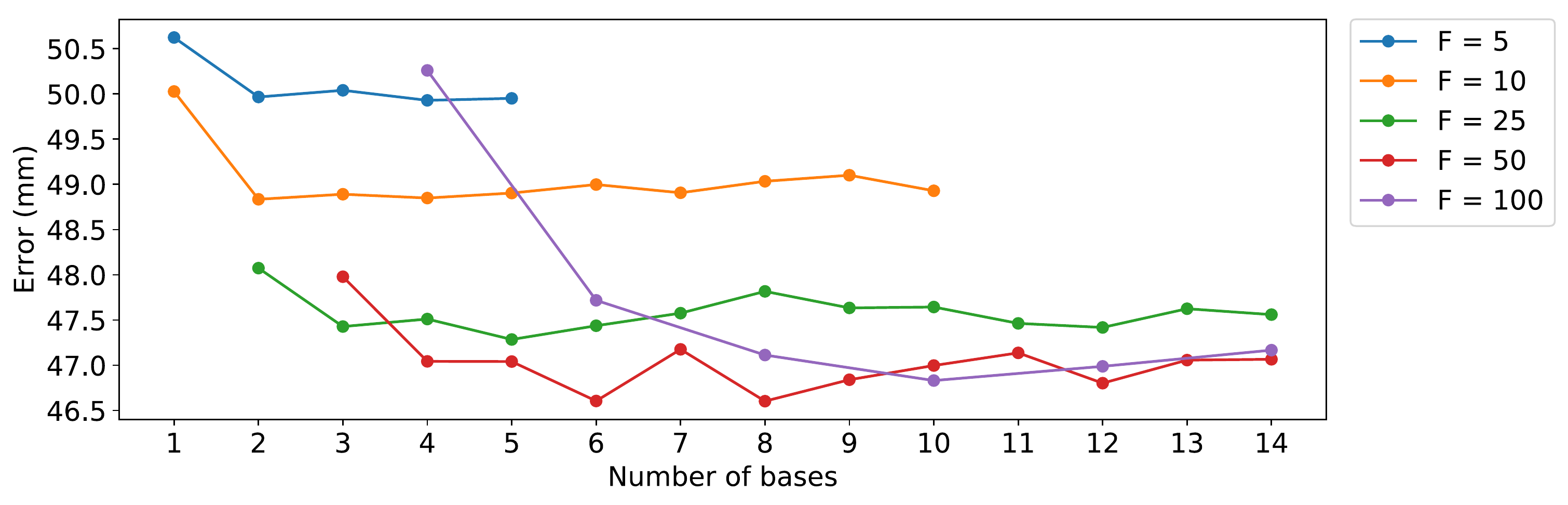}\label{fig:frames-vs-bases}}
\caption{\textbf{(a)}: Average per frame error within a sequence of different sequence lengths. \textbf{(b)}: Estimation error on Human3.6M for different numbers of frames and bases.}
\end{figure*}

\paragraph{Results.} Table \ref{tab:protocol1} shows the results on Human3.6M under protocol 1. We report the performance of our model for input length $F=10$ (with $K=2$ DCT bases), $F=25$ (with $K=5$ DCT bases), and $F=50$ (with $K=8$ DCT bases) corresponding to motion of 0.2s, 0.5s and 1s respectively. Our approach with $F=50$ outperforms all previous single-frame (top half of the table) and temporal-based (bottom half of the table) approaches. Table \ref{tab:protocol2} further shows the results after rigid alignment with the ground truth. Under this protocol, we also show results that are on par with the existing state-of-the-art \cite{pavllo20183d,dabral2018learning}. It is interesting to note the significant improvements of non RNN-based frameworks (\cite{pavllo20183d, dabral2018learning} and ours) over the RNN-based framework \cite{hossain2018exploiting}. Nonetheless, \cite{pavllo20183d, dabral2018learning} estimate only one output frame from a sequence of input frames, while our approach generates stable 3d pose estimates for every frame of an input sequence.
We report the average error of each frame in an input length of $F=\{10, 25, 50\}$ frames in Figure \ref{fig:sequence}. Here, we set the video length to be the same as our network input length, i.e., $L=F$.  
The drop in performance for the frames near both ends are probably due to the small number of bases used in our network.
Overall, our framework is able to generate stable estimation for majority of the frames even for longer sequences.

\begin{table*}[t]
    \centering
    \subfloat[]{
        \scalebox{0.6}{
            \begin{tabular}[t]{l|c||l|c}
                \hline
                GT 2d & MPJPE & GT 2d & MPJPE \\
                \hline\hline
                Martinez \etal \cite{martinez2017simple} & 45.5 & Pavllo \etal \cite{pavllo20183d} $\dagger$ & 37.2\\
                Zhao \etal \cite{zhao2019semantic} & 43.8 & Ours (DCT) ($F=10$) $\dagger$ & 34.4\\
                Lee \etal \cite{lee2018propagating} ($F=1$) & 40.9 & Ours (DCT) ($F=25$) $\dagger$ & 33.0\\
                Hossain \& Little \cite{hossain2018exploiting} $\dagger$ & 39.2 & Ours (DCT) ($F=50$) $\dagger$ & \textbf{32.8} \\
                Lee \etal \cite{lee2018propagating} ($F=3$) $\dagger$ & 38.4 & Ours (SVD) ($F=50$) $\dagger$ & 32.9\\
                \hline
            \end{tabular}
        }
        \label{tab:gt}
    }
    \quad
    \subfloat[]{
        \scalebox{0.6}{
            \begin{tabular}[t]{l|ccc}
                \hline
                MPI-INF-3DHP & PCK & AUC & MPJPE \\
                \hline \hline
                Mehta \etal \cite{mehta2016monocular} & 75.7 & 39.3 & - \\
                Mehta \etal \cite{mehta2017vnect} (ResNet 50) $\dagger$ & 77.8 & 41.0 & - \\
                Mehta \etal \cite{mehta2017vnect} (ResNet 100) $\dagger$ & 79.4 & 41.6 & - \\
                Ours ($F=25$) $\dagger$ & \textbf{83.6} & \textbf{51.4} & \textbf{79.8} \\
                Ours ($F=50$) $\dagger$ & 82.4 & 49.6 & 81.9\\
                \hline
            \end{tabular}
        }
        \label{tab:mpi}
    }
    \caption{\textbf{(a)}: Results on Human3.6M under Protocol 1 using ground truth 2d input. \textbf{(b)}: Results on MPI-INF-3DHP using ground truth 2d input. $\dagger$ indicates methods using temporal information (including ours). Bold-faced numbers indicate best results.}
\end{table*}

\begin{figure*}[t]
\subfloat{\includegraphics[width=\linewidth]{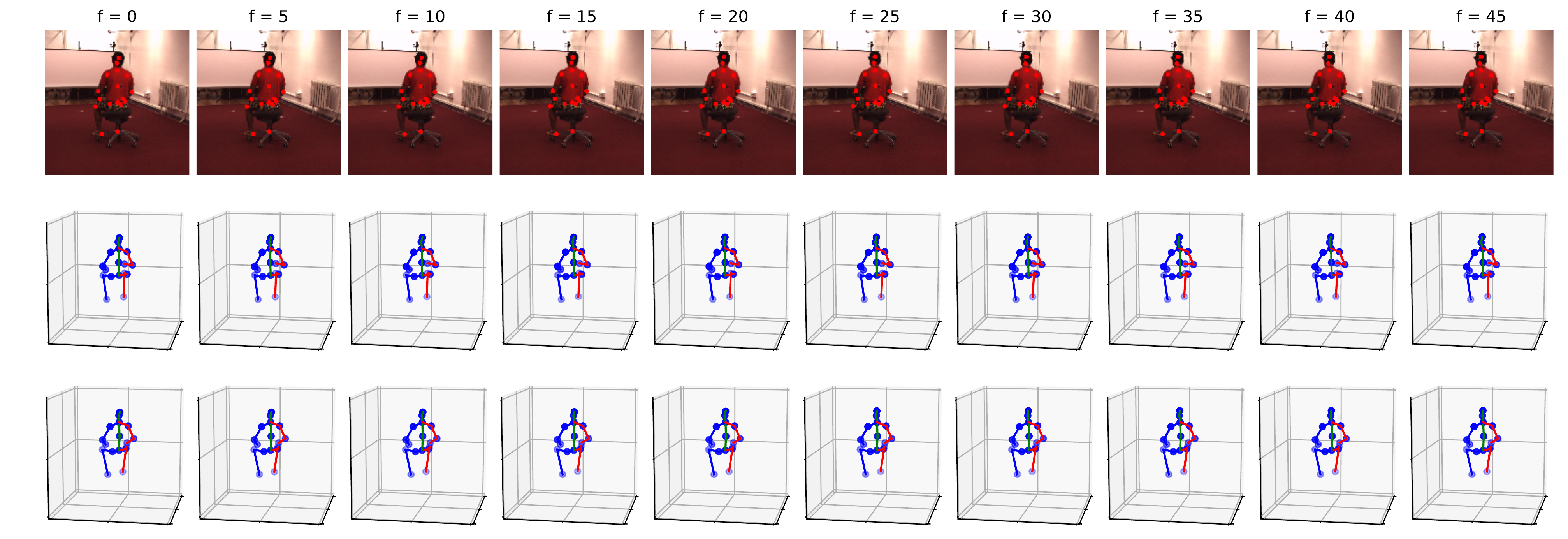}}\\
\subfloat{\includegraphics[width=0.5\linewidth]{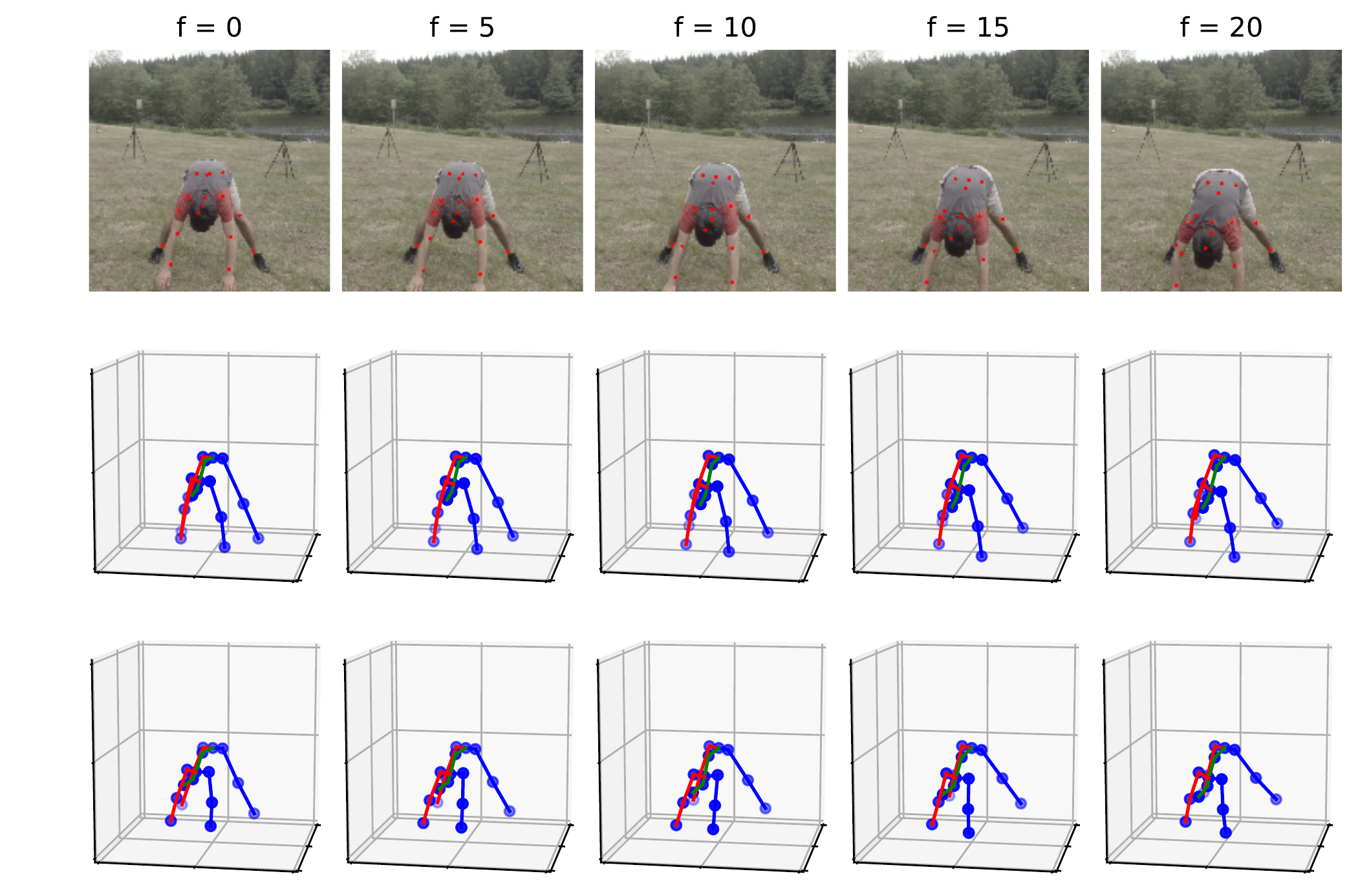}}
\subfloat{\includegraphics[width=0.5\linewidth]{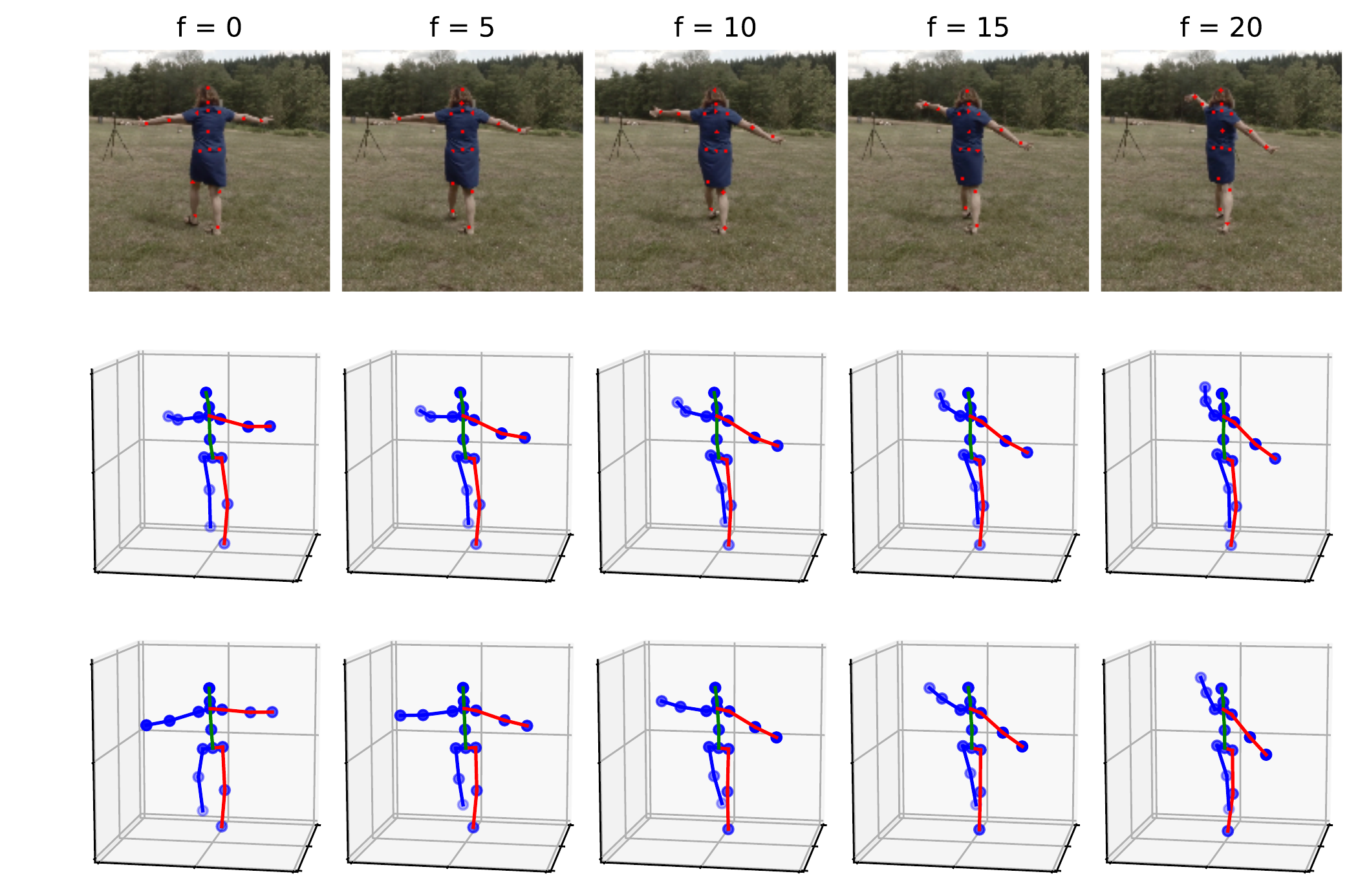}}
\caption{Qualitative results for both indoor and outdoor videos. First row are 2d inputs. Second row are estimated 3d poses. Third row are ground truth 3d poses.}
\label{fig:seq-outdoor}
\end{figure*}

We further evaluate our approach with ground truth 2d poses on Human3.6M as the input. As shown in Table \ref{tab:gt}, our approach significantly improves previous best result by 4.4mm (11.8\%). This demonstrates that regression in the trajectory space is effective in generating highly accurate estimation even though the framework concurrently estimates the 3d poses for a long sequence of frames.
We also report the comparison of using SVD and DCT bases in Table \ref{tab:gt}. Although the SVD bases are computed on a subset of the dataset, it achieves similar result as DCT bases. This suggests that the model is not restricted to any specific bases. We leave the exploration on other bases as future work.
To demonstrate the generalizability of our framework, we evaluate the performance of our model on MPI-INF-3DHP in Table \ref{tab:mpi} with all previous works using ground truth 2d locations as input. $F=25$ frames (with $K=6$ DCT bases) and $F=50$ frames (with $K=8$ DCT bases) are reported for our model. Both settings outperform previous best results under all metrics. Qualitative visualizations are shown in Figure \ref{fig:seq-outdoor}.

We also conduct experiments on how the number of input frames $F$ and bases $K$ affect the performance of the model. Estimation errors on Human3.6M for various $F$ and $K$ settings are shown in Figure \ref{fig:frames-vs-bases}.
We show the plots of $F=5$ at $K\leq 5$ and $F=10$ at $K\leq 10$ because the number of bases cannot be more than the number of frames, i.e., $K \leq F$. The plots for $F = \{25,50,100\}$ are truncated at $K=14$ since the errors stabilized approximately after $K=8$. We can also see that errors converged to $\sim47$mm for $F = \{25,50,100\}$, where $K \geq 8$. This suggests that our network design is versatile and outputs consistent results over different length of input frames.
Although we empirically show the results of our network over $F = \{10, 25, 50\}$ in all the previous results on Human3.6M, the results in Figure \ref{fig:frames-vs-bases} demonstrate that our network is not limited to any fixed number of input frames.


\section{Conclusion}
\label{sec:conclusion}
We proposed a deep learning-based framework that utilizes matrix factorization for sequential 3d human poses estimation. In particular, we showed that the 3d poses in all frames can be represented as a motion matrix factorized into a small number of trajectory bases, and a set of trajectory coefficients. We turned the problem of sequential 3d pose estimation into training a deep network to regress the set of trajectory coefficients from all the input frames. The effectiveness of our framework is demonstrated by achieving state-of-the-art performances on multiple benchmark datasets.


\end{document}